\documentclass[final]{cvpr}

\usepackage{times}
\usepackage{epsfig}
\usepackage{graphicx}
\usepackage{amsmath}
\usepackage{amssymb}
\usepackage{multirow}
\usepackage{color}
\usepackage{subfiles}
\usepackage{amsmath,bm}
\usepackage[pagebackref=true,breaklinks=true,colorlinks,bookmarks=false]{hyperref}

\allowdisplaybreaks[4]

\begin{document}

\title{Anti-aliasing Semantic Reconstruction for Few-Shot Semantic Segmentation}

\author{Binghao Liu$^{1}$ \and Yao Ding$^{1}$ \and Jianbin Jiao$^1$ \and Xiangyang Ji$^2$ \and Qixiang Ye$^{1}$\thanks{Corresponding Author.}
\and PriSDL, EECE, University of Chinese Academy of Sciences$^1$\and Department of Automation, Tsinghua University$^2$

\and\tt\small \{liubinghao18, dingyao16\}@mails.ucas.ac.cn\and\tt\small xyji@tsinghua.edu.cn\and\tt\small \{jiaojb, qxye\}@ucas.ac.cn
}

\maketitle

\pagestyle{empty}  
\thispagestyle{empty} 
\begin{abstract}
Encouraging progress in few-shot semantic segmentation has been made by leveraging features learned upon base classes with sufficient training data to represent novel classes with few-shot examples.
However, this feature sharing mechanism inevitably causes semantic aliasing between novel classes when they have similar compositions of semantic concepts.
In this paper, we reformulate few-shot segmentation as a semantic reconstruction problem, and convert base class features into a series of basis vectors which span a class-level semantic space for novel class reconstruction.
By introducing contrastive loss, we maximize the orthogonality of basis vectors while minimizing semantic aliasing between classes.
Within the reconstructed representation space, we further suppress interference from other classes by projecting query features to the support vector for precise semantic activation.
Our proposed approach, referred to as anti-aliasing semantic reconstruction (ASR), provides a systematic yet interpretable solution for few-shot learning problems.
Extensive experiments on PASCAL VOC and MS COCO datasets show that ASR achieves strong results compared with the prior works.
Code will be released at \href{https://github.com/Bibkiller/ASR}{\color{magenta}github.com/Bibkiller/ASR}.

\end{abstract}

\section{Introduction}
    Over the past few years, we have witnessed the substantial progress of object detection and semantic segmentation~\cite{FreeAnchor2019,LTM2021,UNet,PSPNet,DeepLabV1,MaskRCNN-PAMI2020}. This can be attributed to convolutional neural networks (CNNs) with excellent representation capability and the availability of large datasets with concise mask annotations, especially. However, annotating a large number of object masks is expensive and infeasible in some scenarios ($e.g.$, computer-aided diagnosis systems). Few-shot semantic segmentation, which aims to generalize a model pre-trained on base classes of sufficient data to novel classes with only a few examples, has emerged as a promising technique. 
    
    \begin{figure}[t]
        \centering
        \includegraphics[width=1\linewidth]{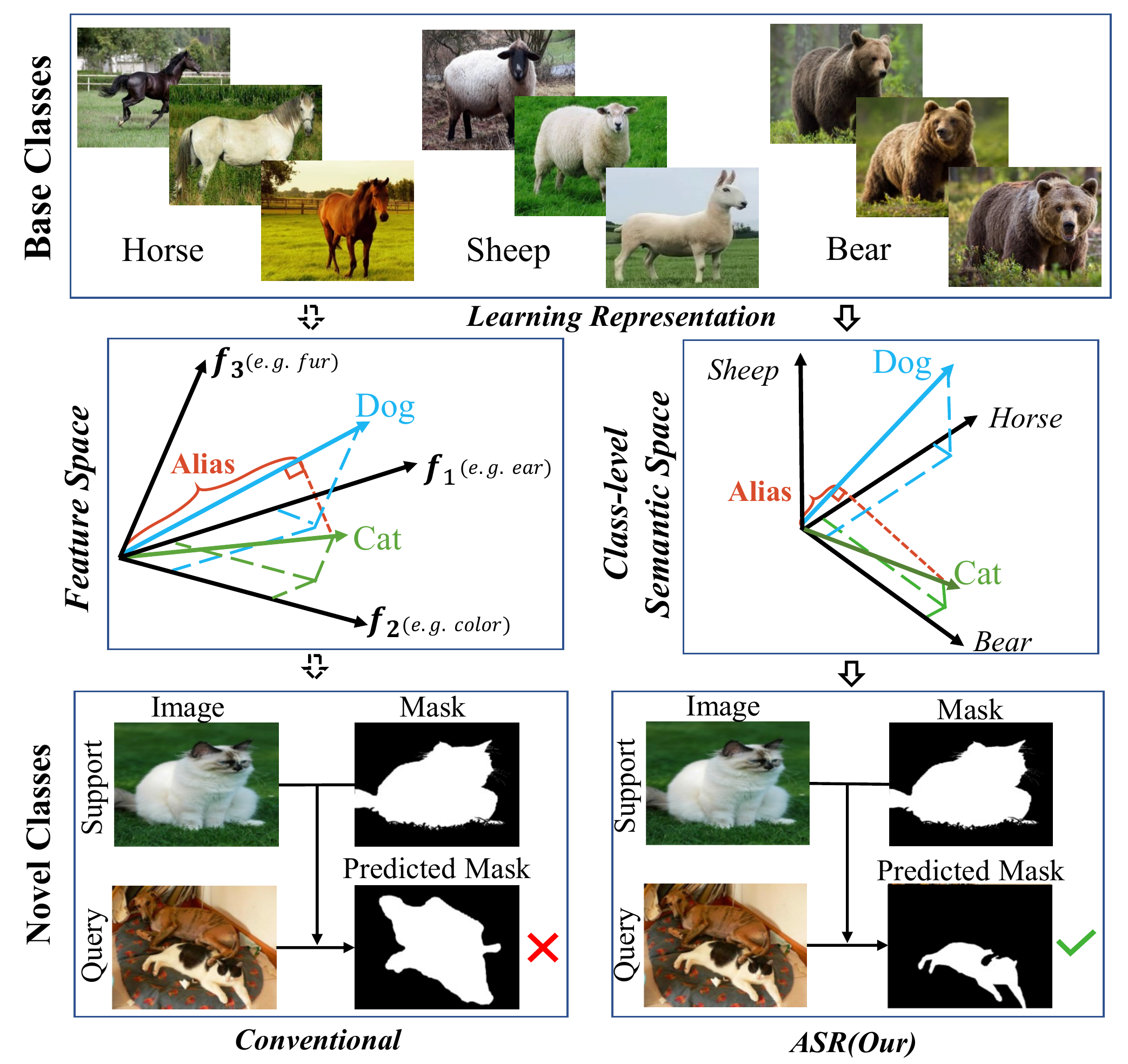}
        \caption{Comparison of conventional methods and our ASR method. While conventional methods represent novel classes ($e.g.$, cat and dog) within the feature space specified for base classes without considering the semantic aliasing, ASR implements semantic reconstruction by constructing a class-level semantic space where basis vectors are orthogonal and the semantic interference is reduced.}
        \label{fig:motivation}
    \end{figure}

    In few-shot segmentation, the generalization process is to utilize features learned upon base classes with sufficient training data to represent novel classes. However, for the overlapped semantics among features, the intricate many-to-many correspondence between features and classes inevitably causes semantic aliasing\footnote{Semantic aliasing refers to an effect that causes classes to be indistinguishable due to the sharing of semantics among features.} between novel classes when they have similar compositions of semantic concepts. For example, a cat and a dog appear in the same query image are confused because they correspond to the similar features of the base classes for bears and sheep, which results in false segmentation, Fig.~\ref{fig:motivation}(left).  

    In this paper, we reformulate the few-shot segmentation task as a semantic reconstruction problem and propose an anti-aliasing semantic reconstruction (ASR) approach. To fulfil semantic reconstruction, we first span a class-level semantic space. During the training phase, convolutional feature channels are categorized into channel groups, each of which is optimized for constructing a basis vector corresponding to a base class. This suppresses the semantic overlap between feature channels. We further introduce a contrastive loss to enhance the orthogonality of basis vectors and improve their representation capability. In the space, the semantic vectors of novel classes are represented by weighted basis vector reconstruction. Due to the potential class-level semantic similarity, the novel class will be reconstructed by its semantic-proximal base classes. In this way, novel classes inherit the orthogonality of base classes and are distinguishable, Fig.~\ref{fig:motivation}(middle right).
  
\begin{figure*}[t]
        \centering 
        \includegraphics[width=1.0\linewidth]{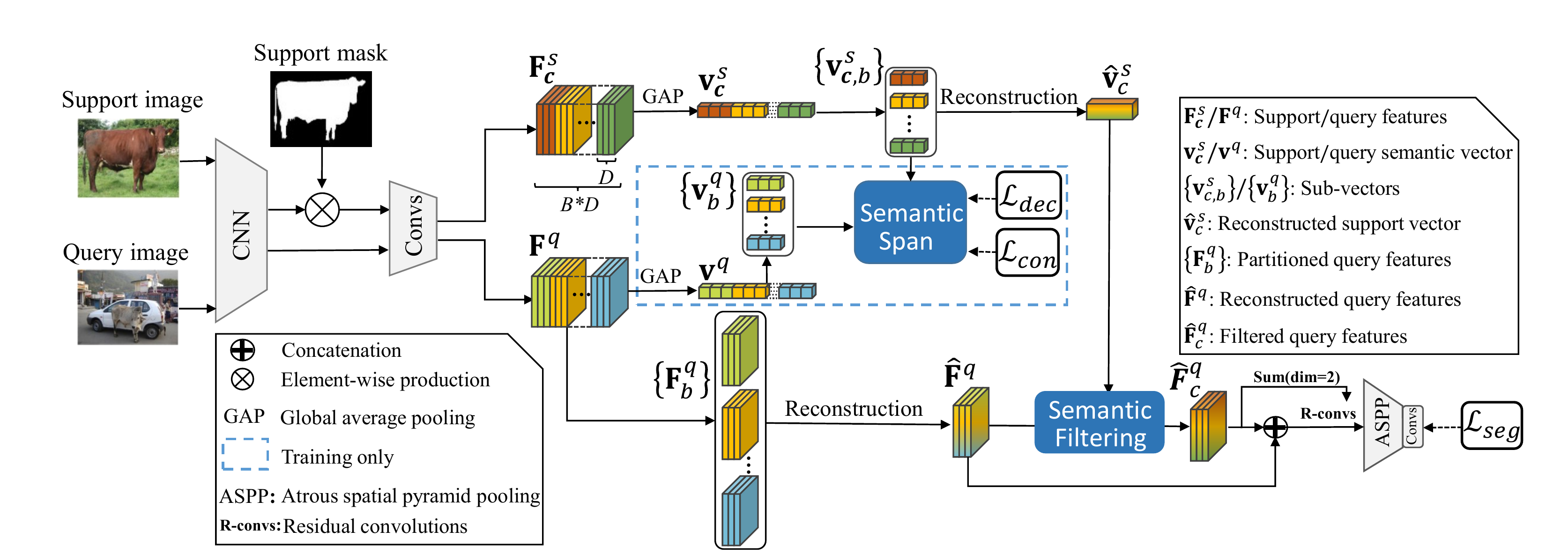}
        \caption{Few-shot segmentation flowchart with Anti-aliasing Semantic Reconstruction (ASR). The flowchart defines a metric learning framework consisting of a support branch (upper) and a query branch (lower), where reconstruction, semantic span, and semantic filtering modules are plugged. While semantic span reduces the semantic aliasing between base classes driven by contrastive loss, semantic filtering aims to suppress interfering semantics within the query image. (Best viewed in color)}
        \label{fig:FrameWork}
    \end{figure*}

    To suppress interfering semantics from the background or other classes within the same query image, we further propose the semantic filtering module, which projects query feature vectors to the reconstructed support vector. As the support images have precise semantics guided by the ground-truth annotations, the projection operation divorces interfering semantics, which facilities the activation of target object classes, Fig.~\ref{fig:motivation}(bottom right). In the metric learning framework, ASR implements semantic anti-aliasing between novel classes and within query images, providing a systematic solution for few-shot learning, Fig.\ \ref{fig:FrameWork}. Such anti-aliasing can be analyzed from perspectives of vector orthogonality and sparse reconstruction, making ASR an interpretable approach.
   
    The contributions of this study include:
    \begin{itemize}
        \item We propose a systematic and interpretable anti-aliasing semantic reconstruction (ASR) approach for few-shot semantic segmentation, by converting the base class features into a series of basis vectors for semantic reconstruction.

        \item We propose semantic span, which reduces the semantic aliasing between base classes for precise novel class reconstruction. Based on semantic span, we further propose semantic filtering, to eliminate interfering semantics within the query image.

        \item ASR improves the prior approaches with significant margins when applied to commonly used datasets. It also achieves good performance under the two-way few-shot segmentation settings.
    \end{itemize}

\section{Related Works}

    \textbf{Semantic Segmentation.} 
        Benefiting from the superiority of fully convolutional networks, semantic segmentation~\cite{DeepLab,TIP-MMM,PSPNet} has progressed substantially in recent years. Relevant research has also provided some fundamental techniques, such as multi-scale feature aggregation~\cite{PSPNet} and atrous spatial pyramid pooling (ASPP)~\cite{DeepLab}, which enhance few-shot semantic segmentation. However, these methods generally require large amounts of pixel-level annotations, which hinders their application in many real-world scenarios.
   
    \textbf{Few-shot Learning.}
       While meta-learning~\cite{LearningToLearn16,Optimization17,MAML17,TaskAgnosticMeta19,CEC,MIAL2021,HFA} contributed important optimization methods and data augmentation~\cite{Hallucinating17,Imaginary18} aggregated performance, metric learning~\cite{MatchNetwork16,Compare2018,CollectSelect19,deng2020labels} with prototype models~\cite{prototype_relation,Chu_2019_CVPR,SRF,DeepEMD,emdv2,Li_2021_CVPR} represent the majority of few-shot learning approaches. In metric learning frameworks, prototypical models convert spatial semantic information of objects to convolutional channels. With prototypes, metric algorithms aim to obtain a high similarity score for similar sample pairs while a low similarity score for dissimilar pairs. For example, Ref.~\cite{CloserLook19} replaced the fully connected layer with cosine similarity. Ref.~\cite{DFV} devises a few-shot visual learning system that performs well on both base and novel classes. DeepEMD~\cite{DeepEMD} proposed the structural distance between dense image representations. Extra margin constraints~\cite{negmargin, li2020boosting} are absorbed into metric learning to further adjust the inter-class diversity and intra-class variance. Despite the popularity of metric learning, the semantic aliasing issue caused by the feature sharing mechanism is unfortunately ignored.

    \textbf{Few-shot Segmentation.}
        Early methods generally utilized a parametric module, which uses features learned through support image(s) to segment the query image. In~\cite{co-FCN} support features were concatenated with the query image to activate features within object regions for segmentation. PGNet~\cite{PGNet} and DAN~\cite{DAN} tackled semantic segmentation with graphs and used graph reasoning to propagate label information to the query image. 

        Following few-shot classification, prototype vectors have been used as semantic representation across feature channels. In~\cite{SG-One}, masked average pooling was utilized to squeeze foreground information within the support image(s) to prototype vectors. CANet~\cite{CaNet} consisted of a two-branch model which performs feature comparison between the support image(s) and the query image guided by prototypes.
        PANet~\cite{PANet} offered highly representative prototypes for each semantic class and performs segmentation over the query image based on pixel-wise matching.
        CRNet~\cite{CRNet} proposed a cross-reference mechanism to concurrently make predictions for both the support image(s) and the query image, enforcing co-occurrence of objects and thereby improving the semantic transfer.
        
        PMMs~\cite{PMMs} and PPNet~\cite{PPN} proposed to decompose objects into parts and represent such parts with mixed prototype vectors to counter semantic mixing. Despite the aforementioned progress, existing methods remain ignorant of the semantic aliasing issue, which causes false (or missing) segmentation of object parts.  SST~\cite{SST} and SimProp~\cite{SimProp} respectively introduced self-supervised finetuning and similarity propagation, which leverage the category-specific semantic constraints to reduce semantic aliasing. However, without considering the orthogonality of base class features, they remain challenged by the semantic aliasing issue.

\section{The Proposed Method}
        \subsection{Problem Definition}
        Few-shot semantic segmentation aims to learn a model ($e.g.$, a network) which can generalize to previously unseen classes. Given two image sets $\mathcal{D}_{base}$ and $\mathcal{D}_{novel}$, classes in $\mathcal{D}_{novel}$ do not appear in $\mathcal{D}_{base}$, it requires to train the feature representation on $\mathcal{D}_{base}$ (which has sufficient data) and test on $\mathcal{D}_{novel}$ (which has only a few annotations). Both $\mathcal{D}_{base}$ and $\mathcal{D}_{novel}$ contain several episodes, each of which consists of a support set ${(\mathbf{A}_i^s,\mathbf{M}_i^s)}_{i=1}^{K}$ and a query set $(\mathbf{A}^q,\mathbf{M}^q)$, where $K,\mathbf{A}_i^s, \mathbf{M}_i^s, \mathbf{A}^q$ and $\mathbf{M}^q$ respectively represent the shot number, the support image, the support mask, the query image, and the query mask. For each training episode, the model is optimized to segment $\mathbf{A}^q$ driven by the segmentation loss $\mathcal{L}_{seg}$. Segmentation performance is evaluated on $\mathcal{D}_{novel}$ across all the test episodes. 

        \subsection{Semantic Reconstruction Framework}
        We propose a semantic reconstruction framework, where the semantics of novel classes are explicitly reconstructed by those of base classes, Fig.~\ref{fig:FrameWork}. Given support and query images, after extracting convolutional features through a CNN, the ground-truth mask is multiplied with support features in a pixel-wised fashion to filter out background features~\cite{SG-One,CaNet,CRNet,PMMs,PPN}. With a convolutional block we reduce the number of feature channels and obtain support features $\mathbf{F}_c^s\in{\mathbb{R}^{H\times W\times (B\times D)}}$ and query features $\mathbf{F}^q\in{\mathbb{R}^{H\times W\times (B\times D)}}$, where $H\times W$, $B$, and $D$ respectively denote the size of feature maps, base class number, and feature channel number. During training phase, $c$ denotes the base class. And $c$ denotes the novel class during testing phase. The convolutional block consists of pyramid convolution layers, which captures features from coarse to fine. To explicitly encode class-related semantics, we averagely partition the feature channels to $B$ groups, corresponding to $B$ base classes. The grouped features $\mathbf{F}_c^s$ and $\mathbf{F}^q$ are further spatially squeezed into two vectors $\mathbf{v}_c^s$ and $\mathbf{v}^q$, termed semantic vectors, by global average pooling, Fig.~\ref{fig:FrameWork}. 
        
        Corresponding to $B$ base classes, the semantic vectors $\mathbf{v}_c^s$ and $\mathbf{v}^q$ consists of $B$ sub-vectors $\{\mathbf{v}^s_{c,b}\}_{\{b=1,2,...,B\}}\in{\mathbb{R}^D}$ and
        $\{\mathbf{v}^q_b\}_{\{b=1,2,...,B\}}\in{\mathbb{R}^D}$. 
        During the training phase, the sub-vectors are used to construct basis vectors in the $B$-dimensional class-level semantic space by the semantic span module, as explained in Section~\ref{subsection:span}, Fig.~\ref{fig:FrameWork}. In the space, a basis vector ($\mathbf{v}_b$) corresponding to the $b$-th base class is defined as $\mathbf{v}_b = \mathbf{v}^s_{c,b}/||\mathbf{v}^s_{c,b}||=\mathbf{v}^q_b/||\mathbf{v}^q_b||$. In the inference phase, the semantic vector for the $c$-th class in support branch can be linearly reconstructed~\cite{lay2016linear}, as
         \begin{equation}
            \hat{\mathbf{v}}^s_{c} = \sum_{b=1}^B{w_{c,b}^s\cdot \mathbf{v}_b},
            \label{eq:reconst}
        \end{equation}
        where $\hat{\mathbf{v}}^s_{c}$ denotes the reconstructed support semantic vector (reconstructed support vector for short), and $w_{c,b}^s$ is the $b$-th element of the weight vector $\mathbf{w}_c^s = softmax([||\mathbf{v}_{c, 1}^s||, ||\mathbf{v}_{c, 2}^s||, \cdots, ||\mathbf{v}_{c, B}^s||])$. Large $w_{c, b}^s$ indicates that the basis vector, whose corresponding base class have strong similarity with the $c$-th class, contributes much to the reconstruction.
        
        Consistently, given a query image, the corresponding query features $\mathbf{F}^q$ are reconstructed by regarding each location on the feature maps as a feature vector. Each location of feature map is reconstructed as $\hat{\mathbf{F}}^q(x,y) = \sum_{b=1}^B{\mathbf{W}_b^q(x,y)\cdot \mathbf{v}_b}$, where $(x,y)$ denotes the coordinates of pixels on the feature map, and $\mathbf{W}_b^q(x,y)$ is defined as the norm of sub-vector $\mathbf{F}^q_{b}(x,y)$. Considering that the query image contains objects not only belonging to the target class but also other classes, we exploit the semantic filtering module, as illustrated in Section~\ref{sub:filter}, to filter out the interfering components in the reconstructed query features for the $c$-th target class semantic segmentation.
        \begin{figure}[t]
        \centering
        \includegraphics[width=1.0\linewidth]{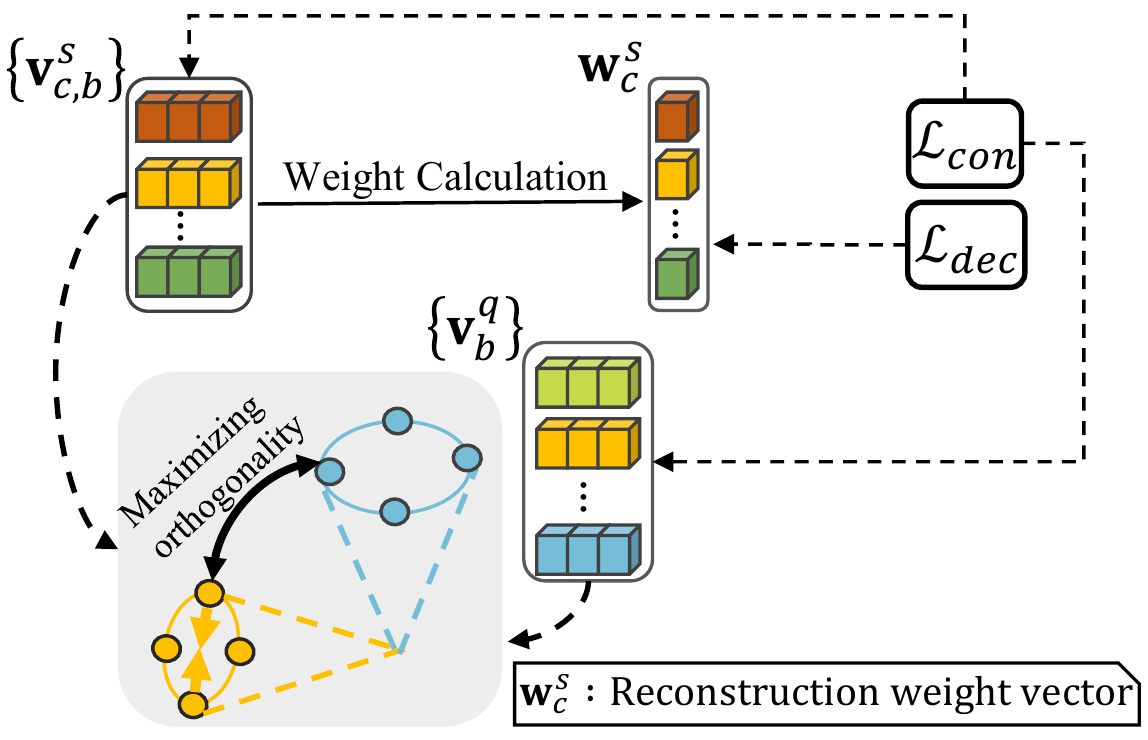}
        \caption{Implementation and illustration of semantic span, which constructs basis vectors and enhances the orthogonality of them. Semantic span is driven by the semantic decoupling loss and constrastive loss. (Best viewed in color)}
        \label{fig:semantic_span}
        \end{figure}

    \subsection{Semantic Span}
    \label{subsection:span}
        Within the origin feature space, when base class features are close to each other, there could be semantic aliasing among novel classes. To minimize semantic aliasing, we propose to span a class-level semantic space in the training phase. To construct a group of basis vectors which tends to be orthogonal and representative, we propose the semantic span module (semantic span for short). As shown in Fig.~\ref{fig:semantic_span}, the semantic span is driven by two loss functions, $i.e.$, semantic decoupling and contrastive losses.
        
        On the one hand, the semantic span targets at constructing basis vectors by regularizing the feature maps so that each group of features is correlated to a special object class. To fulfill this purpose, we propose the following semantic decoupling loss, as
        \begin{equation}
            \mathcal {L}_{dec} = \log (1+e^{-\mathbf{w}_c^s\cdot \mathbf{y}}),
            \label{eq:dec}
        \end{equation}
        where $\mathbf{w}_c^s$ denotes the reconstruction weight vector, and $\mathbf{y} \in \mathbb{R}^B$ denotes the one-hot class label of a support image. Obviously, minimizing $\mathcal{L}_{dec}$ is equivalent to maximize the reconstruction weights related to the specific class ($e.g.$, the $c$-th class), while minimizing those unrelated to it. This defines a soft manner converting a group of features correlated to the semantics of the specific class ($e.g.$, the $c$-th class) to its corresponding basis vector in the class-level semantic space.
        
        On the other hand, the semantic span targets at further enhancing orthogonality of basis vectors, which improve the quality of novel class reconstruction. In details, sub-vectors in $\{\mathbf{v}^s_{c,b}\}_{\{b=1...B\}} \cup \{\mathbf{v}^q_b\}_{\{b=1...B\}}$ belonging to different classes are expected to be orthogonal to each other while those corresponding to the same base classes, $e.g.$, $\mathbf{v}^s_{c,b}$ and $\mathbf{v}^q_{b}$, are expected to have a small vector angle. These two objectives are simultaneously achieved by minimizing the contrastive loss defined as
        \begin{equation}
                \mathcal L_{con} = \frac{e^{1+\sum_{b\neq b'}\lvert cos<\mathbf{v}^s_{c,b},\mathbf{v}^q_{b’}>\rvert}
                }{e^{\lvert cos<\mathbf{v}^s_{c,b},\mathbf{v}^q_b>\rvert}},
        \end{equation}
        where $cos<\cdot>$ denotes the $Cosine$ distance metric of two vectors. In summary, the final loss of the semantic reconstruction framework is defined as:
        \begin{equation}
                \mathcal{L} = \alpha\mathcal{L}_{dec}+\beta\mathcal{L}_{seg}+\gamma\mathcal{L}_{con},
        \end{equation}
        where $\alpha$, $\beta$ and $\gamma$ are weights of the loss functions. Note that $\mathcal{L}_{con}$ is calculated during the later stage of training phase.
        
        \begin{figure}[t]
        \centering
        \includegraphics[width=1.0\linewidth]{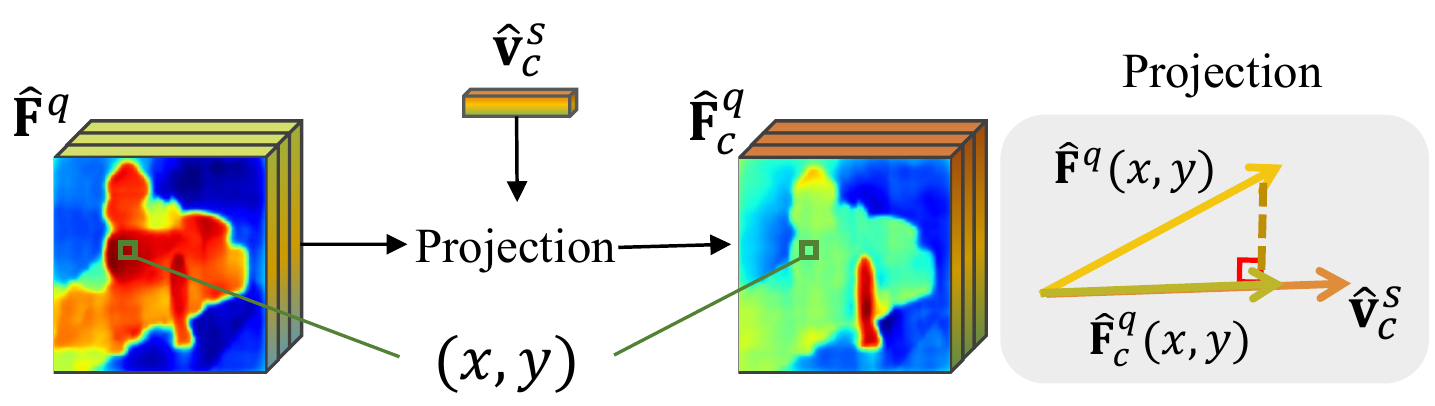}
        \caption{Semantic filtering with vector projection to suppress interfering semantics within the image. (Best viewed in color)}
        \label{fig:filter}
        \end{figure}
        
    \subsection{Semantic Filtering}
    \label{sub:filter}
        When multiple objects from different classes exist in the same query image, the reconstructed features of the query image contains components of all these classes. To pick out objects belonging to the target class and suppress interfering semantics, $i.e.$, divorcing the semantics related to the background or objects from other classes, we propose a semantic filtering module. Moreover, owing to that the reconstructed vectors of different classes are non-collinear, the semantic filtering module is implemented by projecting query feature vectors to the reconstructed support vector, as shown in Fig.~\ref{fig:filter}. This is also based on the fact that the reconstructed support vector has precise semantics because the corresponding features have been multiplied with the ground-truth mask, as shown in Fig.~\ref{fig:FrameWork}. 
        
        On the support branch, we follow Eq.~\ref{eq:reconst} to reconstruct the support features using basis vectors and obtain the reconstructed support vector $\hat{\mathbf{v}}_c^s$. On the query branch, we reconstruct each feature vector $\mathbf{F}^q(x,y)$ on the feature maps in the same way and obtain the reconstructed features $\hat F^q$. We then project $\hat{\mathbf{F}}^q$ to $\hat{\mathbf{v}}_c^s$ to calculate filtered features as
        \begin{equation}
           \hat{\mathbf{F}}_c^q(x,y)= \frac{\hat{\mathbf{F}}^q(x,y)\cdot \hat{\mathbf{v}}_c^s}{\left|\left|\hat{ \mathbf{v}}_c^s\right|\right|} \cdot \frac{\hat{\mathbf{v}}_c^s}{\left|\left|\hat {\mathbf{v}}_c^s\right|\right|},
            \label{eq:projection}
        \end{equation}
        where $(x,y)$ denotes the coordinates of pixels on the feature maps. The intuitive effects of the filter operation are displayed in Fig.~\ref{fig:filter}, which illustrates that the support branch guides the query branch more effectively. The filtered query features $\hat{\mathbf{F}}_c^q$ are further enhanced by a residual convolutional module with iterative refinement optimization and fed to Atrous Spatial Pyramid Pooling (ASPP) to predict the segmentation mask, Fig.~\ref{fig:FrameWork}. For the residual convolutional module, we replace the history mask in CANet~\cite{CaNet} with the squeezed $\hat{\mathbf{F}}_c^q$.
        
        \begin{figure}[t]
        \centering
        \includegraphics[width=1\linewidth]{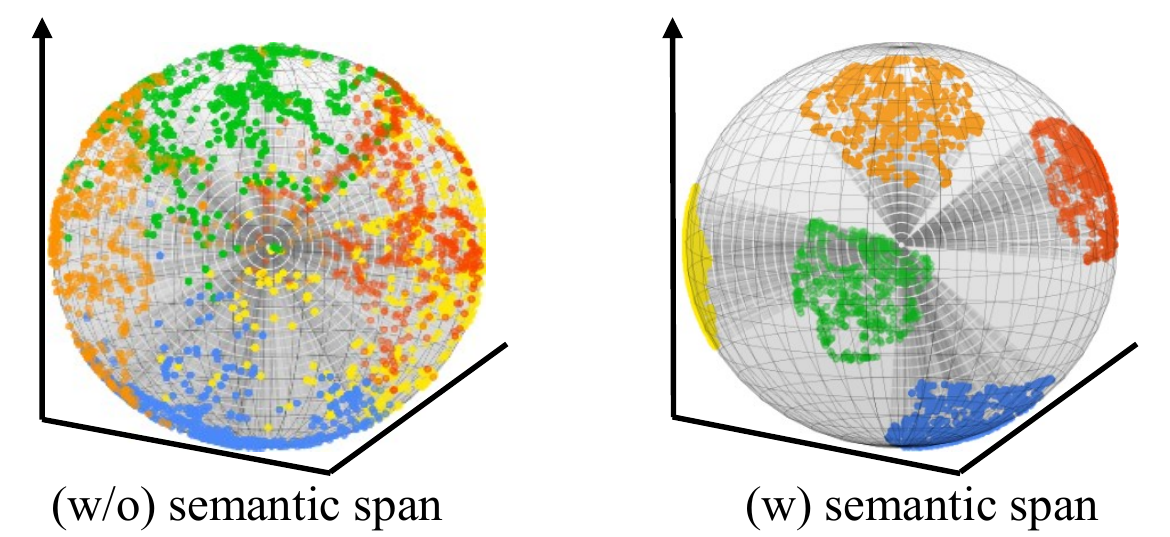}
        \caption{While the semantic vectors without semantic span tends to be mixed up in the semantic space, those with semantic span repel one another towards orthogonality. (Best viewed in color)}
        \label{fig:ortho}
        \end{figure}
    \subsection{Interpretable Analysis}
    \label{sub:analysis}
        ASR can be analyzed from the perspectives of vector orthogonality and sparse reconstruction.
        Without loss of generality, we take the two-dimensional space as an example. Denote ${\mathbf{v_1}, \mathbf{v}_2} \in \mathbb{R}^2$ two unit basis vectors  ($\left \|\mathbf{v}_1  \right \| = 1, \left \|\mathbf{v}_2  \right \| = 1$), which span the space.
        Denote $\theta$ as the angle between $\mathbf{v}_1$ and $\mathbf{v}_2$, and $cos\,\theta = \mathbf{v}_1 \mathbf{v}_2$.
        According to the properties of linear algebra~\cite{lay2016linear}, any vectors, $e.g., {\mathbf{u}_1, \mathbf{u}_2} \in \mathbb{R}^2$, in the spanned space can be linearly reconstructed as
        \begin{equation}
            \left\{\begin{matrix}
            \mathbf{u}_1 = C_1(w_{11}\mathbf{v}_1 + w_{12}\mathbf{v}_2), \forall \mathbf{u}_1 \in \mathbb{R}^2\\ 
            \mathbf{u}_2 = C_2(w_{21}\mathbf{v}_1 + w_{22}\mathbf{v}_2), \forall \mathbf{u}_2 \in \mathbb{R}^2\\
            \end{matrix}\right.
        \end{equation}
        where $w_{11}, w_{12}, w_{21}, w_{22}\in[0,1]$ are reconstruction weights which feed the linear constraints: $ w_{11}+w_{12}=1.0$ and $w_{21}+w_{22}=1.0$. $C_1$ and $C_2$ are scaling constants.
        The cosine similarity between $\mathbf{u}_1$ and $\mathbf{u}_2$ is computed as
        \begin{equation}
        \begin{split}
            &cos<\mathbf{u}_1, \mathbf{u}_2>=\frac{\mathbf{u}_1\mathbf{u}_2}{\left | \mathbf{u}_1 \right | \left | \mathbf{u}_2 \right |} \\
            &= (w_{11}\mathbf{v}_1 + w_{12}\mathbf{v}_2)(w_{21}\mathbf{v}_1 + w_{22}\mathbf{v}_2) \\
            &= w_{11}w_{21} + (w_{11}w_{22}+w_{21}w_{12})\mathbf{v}_1\mathbf{v}_2 + {w_{12}}w_{22} \\
            &= w_{11}w_{21} + (1-w_{11})(1-w_{21}) \\ &\quad\quad\quad\quad+[w_{11}(1-w_{21})+w_{21}(1-w_{11})]cos\,\theta  \\
            &= 1 + (w_{11} + w_{21} - 2w_{11}w_{21})(cos\,\theta- 1),
            \label{eq:analysis}
        \end{split}            
        \end{equation}      
        where $(w_{11} + w_{21} - 2w_{11}w_{21}) \in [0, 1]$ and $(cos\,\theta- 1) \in [-1, 0].$ 
            
        \begin{figure}[t]
        \centering
        \includegraphics[width=1.0\linewidth]{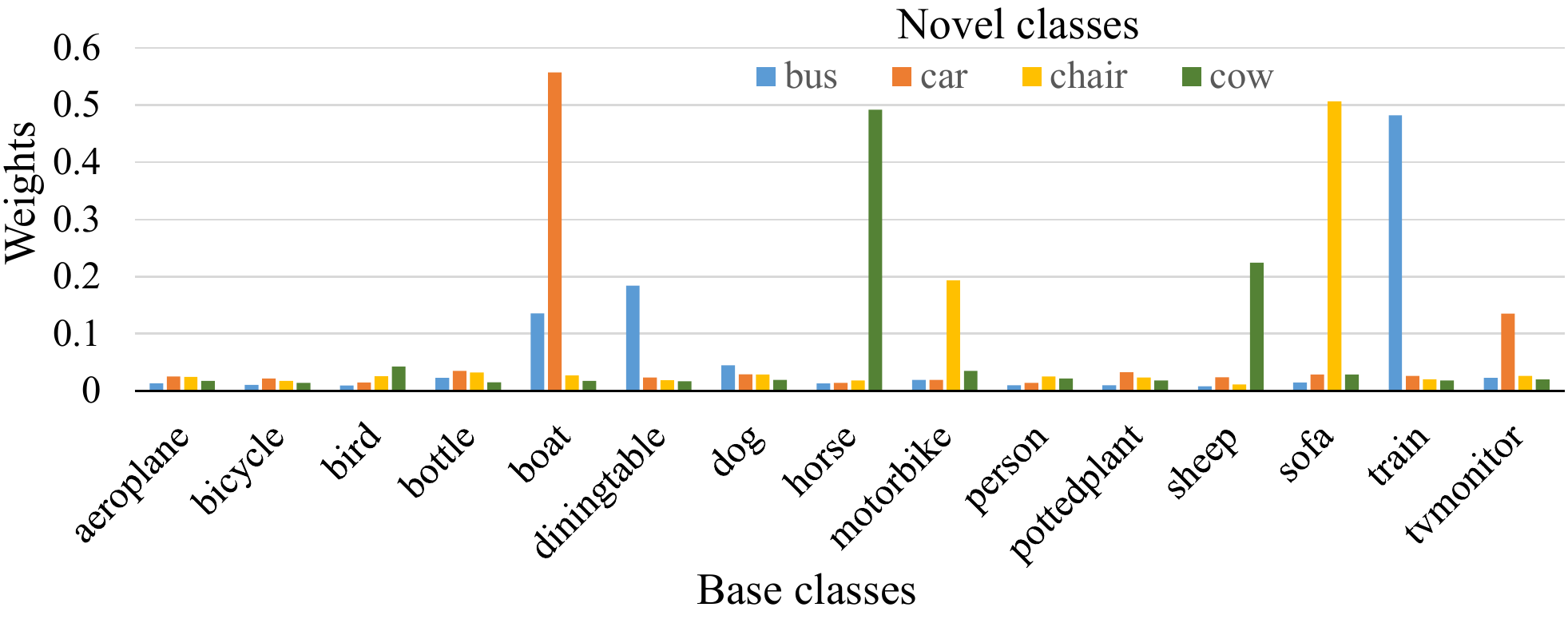}
        \caption{Sparse reconstruction weights of novel classes using the base classes. (Best viewed in color)}
        \label{fig:sparse}
        \end{figure}
        \textbf{Orthogonality.}
        To reduce semantic aliasing of any two novel classes, the angle between their semantic vectors, $\mathbf{u}_1$ and $\mathbf{u}_2$, should be large, known as to reduce $cos<\mathbf{u}_1, \mathbf{u}_2>$. According to the last line of Eq.\ \ref{eq:analysis}, to obtain a small $cos<\mathbf{u}_1, \mathbf{u}_2>$, the term $cos\,\theta$ should approach to $0$, which means that the angle between the basis vectors $\mathbf{v}_1$ and $\mathbf{v}_2$ is large, which implies the orthogonality of basis vectors. The proposed ASR approach satisfies the orthogonality by introducing the semantic span module. As shown in Fig.\ref{fig:ortho}, the statistical visualization results over base classes validate the orthogonality.

        \textbf{Sparse Reconstruction.}
        Refer to the last line of Eq.\ \ref{eq:analysis}, the another manner to reduce the $cos<\mathbf{u}_1, \mathbf{u}_2>$ is enlarging the term $(w_{11} + w_{21} - 2w_{11}w_{21})$. According to the function characteristics, $(w_{11} + w_{21} - 2w_{11}w_{21})$ reaches its maximum when $|w_{11}$-$w_{21}|$ approaches to $1.0$, which contains underlying conditions that $|w_{11}$-$w_{12}|$ and $|w_{21}$-$w_{22}|$ approach to $1.0$ due to the linear constraints. This illustrates that to further aggregate the capability of anti-aliasing and guarantee the discrimination of novel classes, the reconstruction weights for novel classes should be differential and sparse. ASR satisfies these requirements due to the potential class-level semantic similarity according to the statistical results shown in Fig.~\ref{fig:sparse}. Meanwhile, nonzero weights over multiple basis classes other than the dominate one enable ASR to distinguish novel classes from base classes.

 \setlength{\tabcolsep}{4pt}
        \begin{table*}[t]
        \label{table:voc-res}
        \begin{center}
        \resizebox{\textwidth}{35mm}{
        \begin{tabular}{c|c|ccccc|ccccc}
        \hline
        \hline\noalign{\smallskip}
        \multirow{2}{*}{Backbone} & \multirow{2}{*}{Method} & \multicolumn{5}{c|}{1-shot} &\multicolumn{5}{c}{5-shot} \\
        & & Pascal-$5^0$ & Pascal-$5^1$ & Pascal-$5^2$ & Pascal-$5^3$ & Mean & Pascal-$5^0$ & Pascal-$5^1$ & Pascal-$5^2$ & Pascal-$5^3$ & Mean\\
        \noalign{\smallskip}
        \hline
        \noalign{\smallskip}
        \multirow{9}{*}{VGG16} &OSLSM~\cite{OSLSM}&33.60 &55.30 &40.90 &33.50 &40.80 &35.90 &58.10 &42.70 &39.10 &43.95\\
        & co-FCN~\cite{co-FCN}                    &36.70 &50.60 &44.90 &32.40 &41.10&-&-&-&-&-\\
        & SG-One~\cite{SG-One}                    &40.20 &58.40 &48.40 &38.40 &46.30&41.90 &58.60 &48.60 &39.40 &47.10\\
        & PANet~\cite{PANet}                      &42.30 &58.00 &51.10 &41.20 &48.10&51.80 &64.60 &59.80 &46.05 &55.70\\
        & FWB~\cite{FWB-ICCV2019}                 &47.04 &59.64 &52.61 &48.27 &51.90 &50.87 &62.86 &56.48 &50.09 &55.08\\
        &PFENet~\cite{PFENet}                     &56.90 &68.20 &54.40 &52.40 &58.00 &59.00 &69.10 &54.80 &52.90 &59.00\\
        & RPMMs~\cite{PMMs}                       &47.14 &65.82 &50.57 &48.54 &53.02&50.00 &66.46 &51.94 &47.64 &54.01 \\
        & SST~\cite{SST}                          &50.90 &63.00 &53.60 &49.60 &54.30 &52.50 &64.80 &59.50 &51.30 &57.00\\
        &ASR (ours)                                    &49.19 &65.41 &52.58 &51.32 &54.63 &52.52 &66.51 &54.98 &53.85 &56.97\\
        &ASR* (ours)                                    &50.21 &66.35 &54.26 &51.81 &55.66 &53.68 &68.49 &55.03 &54.78 &57.99\\
        \hline\noalign{\smallskip}
        \multirow{8}{*}{Resnet50}   &CANet~\cite{CaNet} &52.50 &65.90 &51.30 &51.90 &55.40 &55.50 &67.80 &51.90 &53.20 &57.10\\
                                    &PGNet~\cite{PGNet} &56.00 &66.90 &50.60 &50.40 &56.00 &57.70 & 68.70 &52.90 &54.60 &58.50 \\
                                    &CRNet~\cite{CRNet} & - & - & - & - & 55.70 & - & - & - & - &58.80 \\

                                    &PPNet~\cite{PPN}     &48.58  &60.58  &\bf55.71  &46.47  &52.84 &58.85 &68.28 &\bf66.77 &\bf57.98 &\bf62.97 \\
                                    &SimPropNet~\cite{SimProp} &54.86 &67.33 &54.52 &52.02 &57.19 &57.20 &68.50 &58.40 &56.05 &60.04 \\
                                    &DAN~\cite{DAN}  &- &- &- &- &57.10 &- &- &- &- &59.50 \\
                                    &PFENet~\cite{PFENet} &\bf61.70 &69.50 &55.40 &\bf56.30 &\bf60.80 &\bf63.10 &70.70 &55.80 &57.90 &61.90\\
                                    &RPMMs~\cite{PMMs}  &55.15 &66.91 &52.61 &50.68 &56.34 &56.28 &67.34 &54.52 &51.00 &57.30 \\
                                    &ASR (ours)  &53.81 & \bf69.56 &51.63 &52.76 &56.94 &56.17 &70.56 &53.89 &53.38 &58.50 \\
                                    &ASR* (ours)  &55.23 & \bf70.36 &53.38 &53.66 &58.16 &59.38 &\bf71.85 &56.87 &55.72 &60.96 \\
        \hline
        \hline
        \end{tabular}}
        \setlength{\tabcolsep}{1.4pt}
        \caption{Mean-IoU performance of 1-way 1-shot and 5-shot segmentation on Pascal-5$^i$. ASR* denotes ASR with multi-scale evaluation.}
        \end{center}
        \end{table*}
        
        \setlength{\tabcolsep}{4pt}
            \begin{table*}[t]
            \begin{center}
            \label{table:coco-res}
            \resizebox{\textwidth}{16mm}{
            \begin{tabular}{c|ccccc|ccccc}
            \hline
            \hline\noalign{\smallskip}
            \multirow{2}{*}{Method} & \multicolumn{5}{c|}{1-shot} &\multicolumn{5}{c}{5-shot}\\
            & COCO-$20^0$ & COCO-$20^1$ & COCO-$20^2$ & COCO-$20^3$ & Mean & COCO-$20^0$ & COCO-$20^1$ & COCO-$20^2$ & COCO-$20^3$ & Mean\\
            \hline
            FWB~\cite{FWB-ICCV2019} &16.98 &17.98 &20.96 &28.85 & 21.19 &19.13 &21.46 &23.93 &30.08 &23.65\\
            PFENet~\cite{PFENet} &\bf34.30 &33.00 &32.30 &30.10 &32.40 &\bf38.50 &38.60 &\bf38.20 &34.30 &\bf37.40\\
            SST~\cite{SST}          &- &- &- &- &22.20 &- &- &- &- &31.30\\
            DAN~\cite{DAN}          &- &- &- &- &24.40 &- &- &- &- &29.60\\
            RPMMs~\cite{PMMs}        &29.53 &\bf36.82 &28.94 &27.02 &30.58 &33.82 &\bf41.96 &32.99 &33.33 &35.52\\
            ASR (ours)                    &29.89 &34.98 & 31.86 & \bf33.51 & \bf32.56 & 31.26 & 37.86 &33.47 & \bf35.21 & 34.35 \\
            ASR* (ours)                    &30.62 & 36.73 & \bf32.68 & \bf35.35 & \bf33.85 & 33.12 & 39.51 &34.16 & \bf36.21 & \bf35.75 \\
            \hline
            \hline
            \end{tabular}}
            \setlength{\tabcolsep}{1.4pt}
            \caption{Mean-IoU performance of 1-shot and 5-shot semantic segmentation on COCO-$20^i$. FWB and PFENet use the ResNet101 backbone while other approaches use the ResNet50 backbone. ASR* denotes ASR with multi-scale evaluation.}
            \end{center}
            \end{table*}
            
\section{Experiments}
    In this section, we first describe the experimental settings. We then report the performance of ASR and compare it with state-of-the-art methods. We finally present ablation studies with experimental analysis and test the effectiveness of ASR on other few-shot learning tasks. 

\subsection{Experimental Settings}
        \textbf{Datasets.} The experiments are conducted on PASCAL VOC 2012~\cite{voc} and MS COCO~\cite{coco} datasets. We combine the PASCAL VOC 2012 with SBD~\cite{SDS11} and separate the combined dataset into four splits. The cross-validation method is used to evaluate the proposed approach by sampling one split as test categories $\mathcal{C}_{test}={4i+1,…,4i+5}$, where $i$ is the index of a split. The remaining three splits are set as base classes for training. The reorganized dataset is termed as Pascal-$5^i$~\cite{DAN,PMMs}.
        Following the settings in~\cite{FWB-ICCV2019,DAN,PMMs} we construct the COCO-$20^i$ dataset. MS COCO is divided into four splits, each of which contains 20 categories. We follow the same scheme for training and evaluation as on the Pascal-$5^i$. The category labels for the four splits are included in the supplementary material. For each split, 1000 pairs of support and query images are randomly selected for performance evaluation.

\setlength{\tabcolsep}{4pt}
\begin{table}[t]
            \centering
            \renewcommand\arraystretch{1.0}
            \begin{tabular}{c|c|c}
                \hline
                \hline
                \noalign{\smallskip}
                Method & 1-shot & 5-shot \\
                \hline
                SG-One~\cite{SG-One}    & 63.9 & 65.9 \\
                PANet~\cite{PANet}      & 66.5 & 70.7 \\
                CANet~\cite{CaNet}      & 66.2 & 69.6 \\
                PGNet~\cite{PGNet}      & 69.9 & 70.5 \\
                CRNet~\cite{CRNet}      & 66.8 & 71.5 \\
                PFENet~\cite{PFENet}    &\bf73.30 &73.90\\
                DAN~\cite{DAN}          &71.90 &72.30\\
                PPNet~\cite{PPN}          &69.19 &\bf75.76\\
                \hline
                ASR (ours)                &71.33 &72.51 \\
                ASR* (ours)               &72.86 &74.12 \\
                \hline
                \hline
            \end{tabular}
            \setlength{\tabcolsep}{1.4pt}
            \caption{Comparison of FB-IoU performance on Pascal-5$^i$. ASR* denotes ASR with multi-scale evaluation.}
            \label{tab:fb-iou} 
        \end{table}

        \textbf{Training and Evaluation.} We use CANet~\cite{CaNet} without attention modules as the baseline. In training, we set the learning rate as 0.00045. The segmentation model (network) is trained for 200000 steps with the poly descent training strategy and the stochastic gradient descent (SGD) optimizer. Several data augmentation strategies including normalization, horizontal flipping, gaussian filtering, random cropping, random rotation and random resizing are used. We adopt both the single-scale and multi-scale~\cite{CaNet,PGNet,CRNet} evaluation strategies during testing. Our approach is implemented upon the PyTorch 1.3 and run on Nvidia Tesla V100 GPUs.
        
        \textbf{Evaluation Metric.} 
        Following~\cite{PANet,FWB-ICCV2019,CaNet}, we use the mean Intersection over Union (mIoU) and binary Intersection over Union (FB-IoU) as the performance evaluation metrics. The mIoU calculates the per-class foreground IoU and averages the IoU for all classes to obtain the final evaluation metric. The FB-IoU calculates the mean of foreground IoU and background IoU over all images regardless of category. For category $k$, IoU is defined as $IoU_k = TP_k/(TP_k+FP_k+FN_k)$, where the $TP_k,FP_k$ and $FN_k$ are the number of true positives, false positives and false negatives in segmentation masks. mIoU is the average of IoUs for all the test categories and FB-IoU is the average of IoUs for all the test categories and the background. We report the segmentation performance by averaging the mIoUs on the four cross-validation splits.
        
        \begin{figure}[t]
        \centering
        \includegraphics[width=1\linewidth]{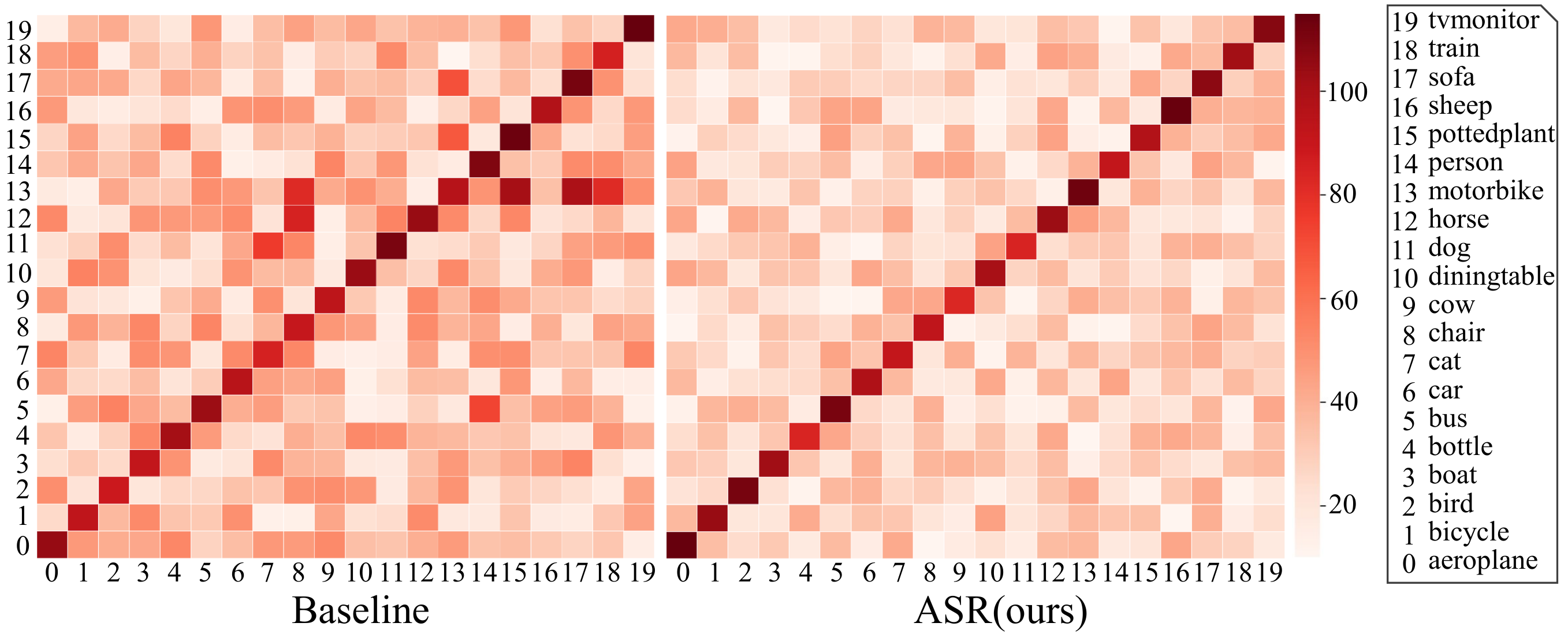}
        \caption{Confusion matrices of baseline method and the proposed ASR approach. (Best viewed in color)}
    \label{fig:confusion}
    \end{figure}
    
    \subsection{Segmentation Performance}
        \textbf{PASCAL VOC.}
        In Table {\color{red}1}, we report the performance on Pascal VOC. 
        ASR outperforms the prior methods with significant margins.
        Under 1-shot settings, with a VGG16 backbone, it respectively outperforms RPMMs~\cite{PMMs} and SST \cite{SST} by 2.64\% and 1.36\%. Under the 1-shot settings, with a ResNet50 backbone, ASR outperforms CANet~\cite{CaNet} and RPMMs~\cite{PMMs} method by 2.76\% and 1.82\%. Under the 5-shot settings, ASR is comparable to the state-of-the-art method. It is worth mentioning that the SST and PPNet used additional $k$-shot fusion strategies while ASR uses a simple averaging strategy to get five-shot results. In Table\ \ref{tab:fb-iou}, ASR is compared with state-of-the-art approaches with respect to FB-IoU. FB-IoU calculates the mean of foreground IoU and background IoU over images regardless of the categories, which reflects how well the full object extent is activated. ASR is on par with the compared methods, if not outperforms.
        
        \textbf{MS COCO.} 
        In Table {\color{red}2}, we report the segmentation performance on MS COCO. ASR outperforms the prior methods in most settings. Particularly under the 1-shot setting, it improves RPMMs~\cite{PMMs} by 3.27\%. Under 5-shot setting, it improves DAN~\cite{DAN} by 6.15\%, which are significant margins.
        For the MS COCO dataset with larger semantic aliasing for the more object categories, semantic reconstruction demonstrated larger advantages. For the larger object category number, we construct a space using more orthogonal basis vectors, which have stronger ability of representation and discrimination.
        According to Section~\ref{sub:analysis}, semantic aliasing among novel classes is suppressed effectively. That is why ASR achieves larger performance gains on the MS COCO dataset.
    
     \begin{figure*}[t]
        \centering
        \includegraphics[width=1.0\linewidth]{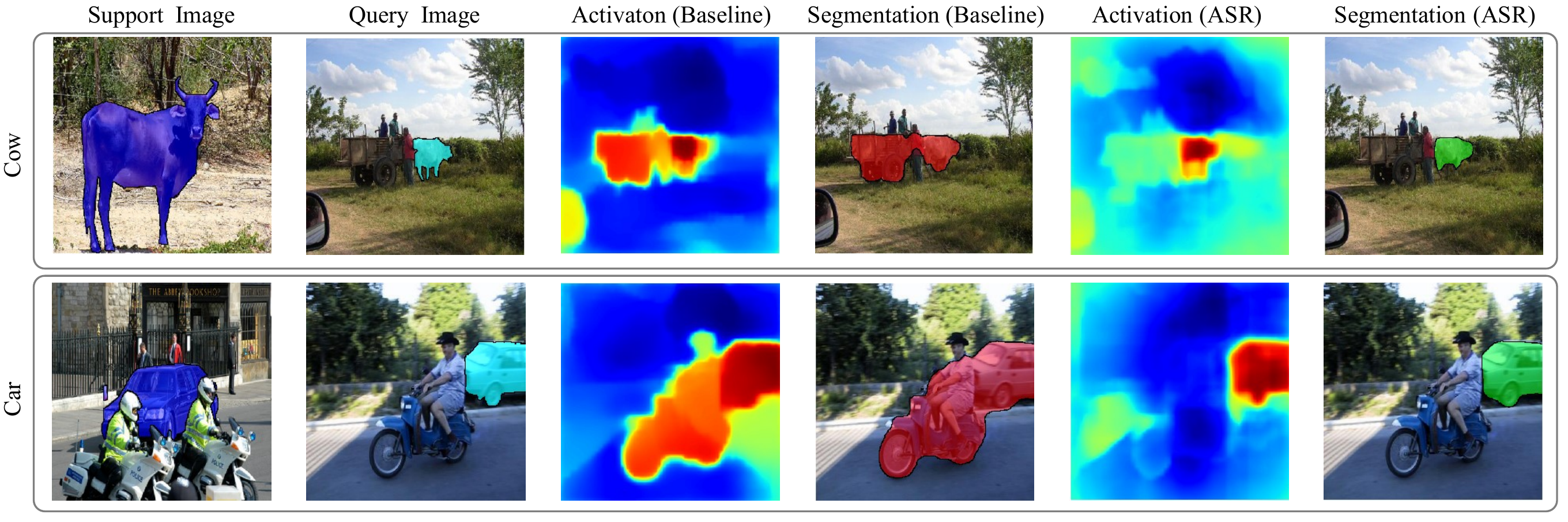}
        \caption{Semantic segmentation results. Compared with the baseline method~\cite{CaNet}, ASR (ours) reduces false positive pixels as well as activating more pixels within target object extent. (Best viewed in color)}
        \label{fig:results}
    \end{figure*}
    
       \subsection{Visualization Analysis}
        
        We sampled 4000 images from 20 classes in PASCAL VOC, and drew the confusion matrix according to the segmentation results, Fig.~\ref{fig:confusion}. ASR effectively reduce semantic aliasing among classes. 
        We further visualize segmentation results and compare them with baseline, Fig.~\ref{fig:results}. 
        Based on the anti-aliasing representation of novel classes and semantic filtering, ASR reduces the false positive segmentation caused by interfering semantics within the query images.
        
        \subsection{Ablation Studies}
        \begin{table}[!t]
        \centering
        \resizebox{0.45\textwidth}{12mm}{
        \begin{tabular}{cccc}
            \hline
            \hline
            \noalign{\smallskip}
             Semantic Reconst. & Semantic Span & Semantic Filter. & mIoU \\
            \hline
            & & & 54.95 \\
             \checkmark &  &  &53.26 \\
             & & \checkmark& 53.12\\
            \checkmark&\checkmark &  &55.98  \\
            \checkmark& \checkmark& \checkmark &\bf 58.64  \\
            \hline
            \hline
        \end{tabular}}
        \setlength{\tabcolsep}{1.4pt}
        \caption{Ablation of ASR modules. The baseline is CANet.}
        \label{tab:ablation} 
    \end{table}
    
    \begin{table}[!t]
        \centering
        \begin{tabular}{c|c|c|c|c}
            \hline
            \hline
            \noalign{\smallskip}
             Concat. & Cosine & Conv. & Projection  & mIoU\\
            \hline
            \checkmark& & & &58.21\\
             & \checkmark & & &57.78 \\
            & & \checkmark & & 58.32 \\
            & & &\checkmark & \bf 58.64 \\
            \hline
            \hline
        \end{tabular}
        \setlength{\tabcolsep}{1.4pt}
        \caption{Comparison of semantic filtering strategies. Concat., Cosine, Conv., Projection denote vector concatenation, cosine similarity, convolutional operation, and vector projection, respectively.}
        \label{tab:Rectify} 
    \end{table}
        
        \textbf{Semantic Span.} 
        In Table~\ref{tab:ablation}, when simply introducing semantic reconstruction to the baseline method, the performance slightly drops. By using the semantic span module, we improved the performance from 53.26\% to 55.98\%, demonstrating the necessity of establishing orthogonal basis vectors during semantic reconstruction.
        
        \textbf{Semantic Filtering.}
         As shown in Table~\ref{tab:ablation}, directly applying semantic filtering on the baseline method harms the performance because the support features contain aliasing semantics. 
         By combining all the modules, ASR improves the mIoU by 2.66\% (58.64\% vs. 55.98\%). In Table~\ref{tab:Rectify}, four filtering strategies are compared. The vector projection strategy defined in Section~\ref{sub:filter} achieves the best result. Vector projection utilizes the characteristic of vector operations to retain semantics related to the target class and suppress unrelated semantics at utmost.
     
\begin{figure}[t]
            \centering
            \includegraphics[width=1.0\linewidth]{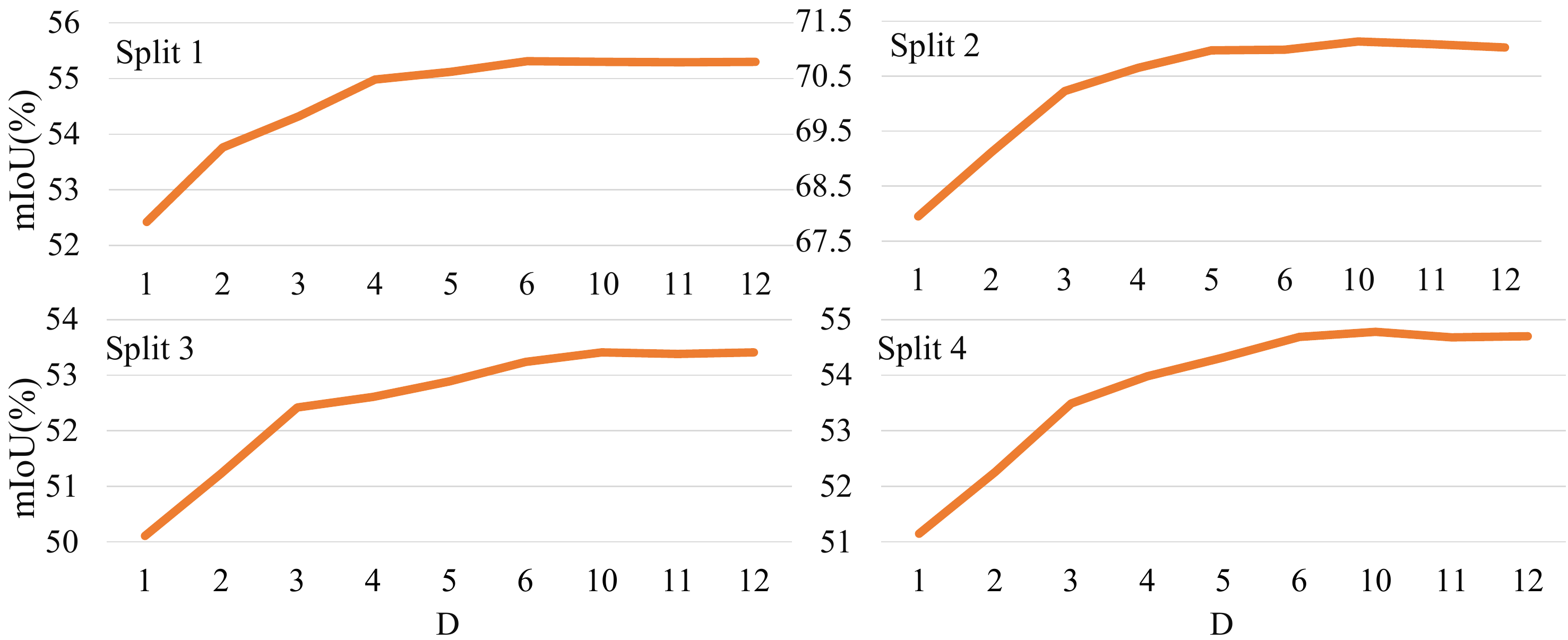}
            \caption{Performance over channel number (D) of basis vectors.}
        \label{fig:line}
        \end{figure}

        \textbf{Channel Number ($D$).}
        The channel number ($D$) of features to construct basis vectors is an important parameter which affects the orthogonality of basis vectors. From Fig.~\ref{fig:line} we can see that the performance improves with the increase of $D$ and starts to plateau when $D=8$, where the orthogonality of different basis vectors is sufficient for novel class reconstruction. For the MS COCO dataset $D$ is set to 30. 

        \textbf{Model Size and Efficiency.}
        The model size of ASR is 36.7M, which is slightly larger than that of the baseline method~\cite{CaNet} (36.3M) but much smaller than other methods, such as OSLSM~\cite{OSLSM} (272.6M) and FWB~\cite{FWB-ICCV2019} (43.0M). With a Nvidia Tesla V100 GPU, the inference speed is 30 FPS, which is comparable with that of CANet (29 FPS).
    
        \subsection{Two-way Few-shot Segmentation}
        Following the settings in~\cite{PPN}, we conduct two-way one-shot segmentation experiments on PASCAL VOC. From Tab.~\ref{tab:2way} one can see that ASR outperforms PPNet~\cite{PPN} with a significant margin (53.13\% vs. 51.65\%). Because two-way segmentation requires not only to segment targets objects but also to distinguish different classes, the model is more sensitive to semantic aliasing. Our ASR approach effectively reduces the semantic aliasing between novel classes and thereby achieves superiors segmentation performance.
        
        \begin{table}[!t]
        \centering
        \resizebox{1.0\linewidth}{8mm}{
        \begin{tabular}{c|c|c|c|c|c}
            \hline
            \hline
            Method & Pascal-$5^0$ &Pascal-$5^1$ &Pascal-$5^2$ &Pascal-$5^3$ & Mean\\
            \hline
            PPNet~\cite{PPN} &47.36 & 58.34 & \bf52.71 & 48.18 & 51.65\\
            ASR (ours) &\bf 49.35  &\bf 60.68  & 52.12  & \bf 50.38 &\bf 53.13 \\
            \hline
            \hline
        \end{tabular}}
        \setlength{\tabcolsep}{4pt}
        \caption{Mean-IoU performance of 2-way 1-shot segmentation on PASCAL VOC.}
        \label{tab:2way}
        \end{table}

\section{Conclusion}
We proposed Anti-aliasing Semantic Reconstruction (ASR), by converting base class features to a series of basis vectors, which span a semantic space. During training, ASR maximized the orthogonality while minimize the semantic aliasing of base classes, which facilities novel class reconstruction. During inference, ASR further suppresses interfering semantics for precise activation of target object areas. On the large-scale MS COCO dataset, ASR improved the performance of few-shot segmentation, in striking contrast with the prior approaches. As a systematic yet interpretable method for semantic representation and semantic anti-aliasing, ASR provides a fresh insight for the few-shot learning problem.

\textbf{Acknowledgement.} This work was supported by Natural Science Foundation of China (NSFC) under Grant 61836012, 61620106005 and 61771447, the Strategic Priority ResearchProgram of Chinese Academy of Sciences under Grant No. XDA27000000, CAAI-Huawei MindSpore Open Fund and MindSpore deep learning computing framework at www.mindspore.cn.

{\small
\bibliographystyle{ieee_fullname}
\bibliography{cvpr}
}

\end{document}